\documentclass[pdflatex,sn-mathphys]{sn-jnl}

\jyear{2021}%

\usepackage{hyperref}
\usepackage[switch]{lineno} 

\begin{document}

\title[Hierarchical Message-Passing Graph Neural Networks]{Hierarchical Message-Passing Graph Neural Networks}


\author[1]{\fnm{Zhiqiang} \sur{Zhong}}\email{zhiqiang.zhong@@uni.lu}

\author*[2]{\fnm{Cheng-Te} \sur{Li}}\email{chengte@mail.ncku.edu.tw}

\author*[1,3]{\fnm{Jun} \sur{Pang}}\email{jun.pang@uni.lu}

\affil[1]{\orgdiv{Faculty of Science, Technology and Medicine}, \orgname{University of Luxembourg}, \orgaddress{\city{Esch-sur-Alzette}, \country{Luxembourg}}}

\affil[2]{\orgdiv{Institute of Data Science and the Department of Statistics}, \orgname{National Cheng Kung University}, \orgaddress{\city{Tainan}, \country{Taiwan}}}

\affil[3]{\orgdiv{Interdisciplinary Centre for Security, Reliability and Trust}, \orgname{University of Luxembourg}, \orgaddress{\city{Esch-sur-Alzette}, \country{Luxembourg}}}



\abstract{
Graph Neural Networks (GNNs) have become a prominent approach to machine learning with graphs and have been increasingly applied in a multitude of domains.
Nevertheless, since most existing GNN models are based on \textit{flat} message-passing mechanisms, two limitations need to be tackled: 
(i) they are costly in encoding long-range information spanning the graph structure; 
(ii) they are failing to encode features in the high-order neighbourhood in the graphs as they only perform information aggregation across the observed edges in the original graph.
To deal with these two issues, we propose a novel \textit{Hierarchical Message-passing Graph Neural Networks} framework. 
The key idea is generating a hierarchical structure that re-organises all nodes in a flat graph into multi-level super graphs, along with innovative intra- and inter-level propagation manners. 
The derived hierarchy creates shortcuts connecting far-away nodes so that informative long-range interactions can be efficiently accessed via message passing and incorporates meso- and macro-level semantics into the learned node representations.
We present the first model to implement this framework, termed \textit{Hierarchical Community-aware Graph Neural Network} (HC-GNN), with the assistance of a hierarchical community detection algorithm.
The theoretical analysis illustrates HC-GNN's remarkable capacity in capturing long-range information without introducing heavy additional computation complexity.
Empirical experiments conducted on $9$ datasets under transductive, inductive, and few-shot settings exhibit that HC-GNN can outperform state-of-the-art GNN models in network analysis tasks, including node classification, link prediction, and community detection.
Moreover, the model analysis further demonstrates HC-GNN's robustness facing graph sparsity and the flexibility in incorporating different GNN encoders. 

}




\maketitle

\section{Introduction}
\label{sec:introduction}
Graphs are a ubiquitous data structure that models objects and their relationships within complex systems, such as social networks, biological networks, recommendation systems, etc~\cite{WPCLZY21}.
Learning node representation from a large graph has been proved as a useful approach for a wide variety of network analysis tasks, 
including link prediction~\cite{ZC18}, node classification~\cite{ZAL18} and community detection~\cite{CLB19}.

Graph Neural Networks (GNNs) are currently one of the most promising paradigms to learn and exploit node representations due to their effective ability to encode node features and graph topology in transductive, inductive, and few-shot settings~\cite{ZCZ20}.
Many existing GNN models follow a similar \textit{flat} message-passing principle where information is iteratively passed between adjacent nodes along observed edges. 
Such a paradigm is able to incorporate local information surrounded by each node~\cite{GSRVD17}.
However, it has been proven to suffer from several drawbacks~\cite{XHLJ19,MWW20,LWWL20}.

Among these deficiencies of flat message-passing GNNs, the limited ability for information aggregation over long-range has attracted significant attention~\cite{LHW18}, since most graph-related tasks require the interactions between nodes that are not directly connected~\cite{AY21}.
That said, flat message-passing GNNs struggle in capturing dependencies between distant node pairs. 
Inspired by the outstanding effectiveness of very deep neural network models has been demonstrated in computer vision and natural language processing domains~\cite{LBH15}, a natural solution is stacking lots of GNN layers together to directly increase the receptive field of each node. 
Consequently, deeper models have been proposed by simplifying the aggregation design of GNNs and accompanied by well-designed normalisation units or specific gradient descent method~\cite{CWHDL20,GCZSG20}.
Nevertheless, Alon and Yahav have theoretically shown that flat GNNs are susceptible to being a \textit{bottleneck} when aggregating messages across a long path and lead to severe \textit{over-squashing} issues~\cite{AY21}. 

\begin{figure*}[!ht]
\centering
\includegraphics[width=1.\linewidth]{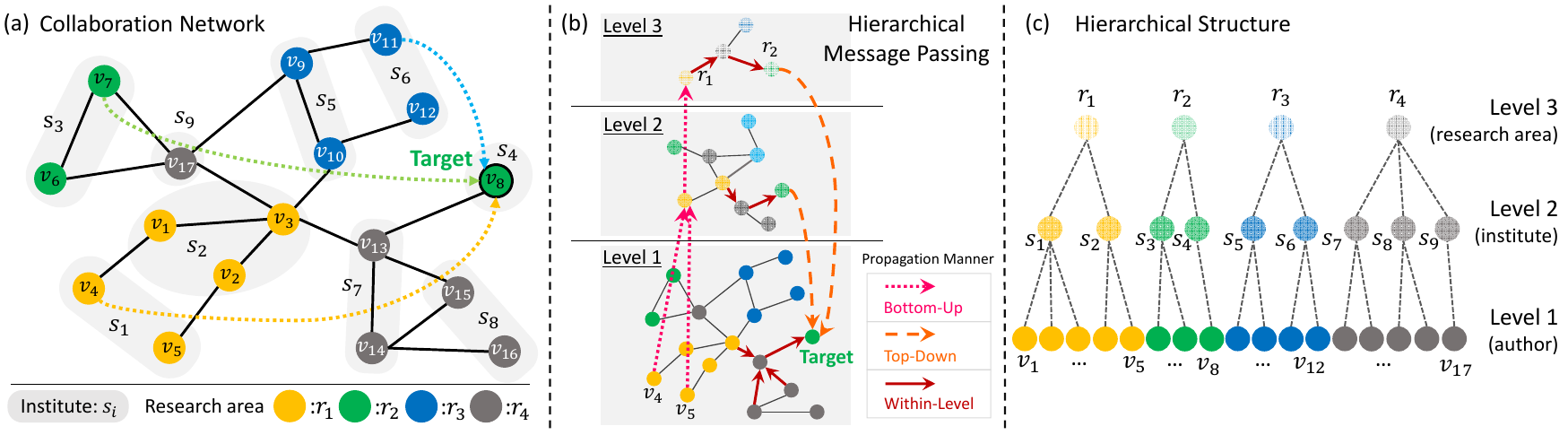}
\caption{Elaboration of the proposed hierarchical message passing:
(a) a collaboration network,
(b) an illustration of hierarchical message-passing mechanism based on (a) and (c), and
(c) an example of the identified hierarchical structure.
}
\label{fig:general_structure}
\end{figure*}

On the other hand, in this paper, we further argue another crucial deficiency of flat message-passing GNNs is that they rely on only aggregating messages across the observed topological structure. 
The hierarchical semantics behind the graph structure provides useful information and should be incorporated into the learning of node representations. 
Taking the collaboration network in Fig.~\ref{fig:general_structure}-(a) as an example;
author nodes highlighted in light yellow come from the same institutes, and nodes filled with different colours indicate authors in various research areas.
In order to generate the node representation of a given author, existing GNNs mainly capture the co-author level information depending on the explicit graph structure.
However, information hidden at \textit{meso} and \textit{macro} levels is neglected. 
In the example of Fig.~\ref{fig:general_structure}, meso-level information means authors belong to the same institutes and their connections to adjacent institutes. 
Macro-level information refers to authors of the same research areas and their relationship with related research areas. 
Both meso- and macro-level knowledge cannot be directly modelled through flat message passing via observed edges.

In this paper, we investigate the idea of a hierarchical message-passing mechanism to enhance the information aggregation pipeline of GNNs. 
The ultimate goal is to make the node representation learning process aware of both long-range interactive information and implicit multi-resolution semantics within the graph.

We note that a few graph pooling approaches have recently delivered various attempts to use the hierarchical structure idea~\cite{GJ19,YYMRHL18,HLLLL19,RST20,LCZT20}.
\textsc{g-U-Net}~\cite{GJ19} and GXN~\cite{LCZT20} employ a bottom-up and top-down pooling operation; however, they do not allow long-range message-passing.
\textsc{DiffPool}~\cite{YYMRHL18}, \textsc{AttPool}~\cite{HLLLL19} and ASAP~\cite{RST20} target at graph classification tasks instead of enabling node representations to capture long-range dependencies and multi-grained semantics of one graph.
Moreover, P-GNNs~\cite{YYL19} create a different information aggregation mechanism that utilises sampled anchor nodes to impose topological position information into learning node representations.
While P-GNNs can capture global information, the hierarchical semantics mentioned above is still overlooked, and the global message-passing is not realised. 
Besides, the anchor-set sampling process is time-consuming for large graphs, and it cannot work well under the inductive setting.

Specifically, we present a novel framework, \textit{Hierarchical Message-passing Graph Neural Networks} (HMGNNs), elaborated in Fig.~\ref{fig:general_structure}.
In detail, HMGNNs can be organised into the following four phases.
\begin{enumerate}
\item[(i)] Hierarchical structure generation.
To overcome long-distance obstacles in the process of GNN message-passing,
we propose to use a hierarchical structure to reduce the size of graph $\mathcal{G}$ gradually,
where nodes at each level $t$ are integrated into different super nodes ($s^{t+1}_{1}, \dots, s^{t+1}_{n}$) at each level $t\!+\!1$.
\item[(ii)] $t$-level super graph construction.
In order to allow the message passing among generated same-level super nodes,
we construct a super graph $\mathcal{G}_t$ based on the connections between nodes at its lower level $t\!-\!1$.
\item[(iii)] Hierarchical message propagation.
With the generated hierarchical structure for a given graph,
we develop three propagation manners, including bottom-up, within-level and top-down.
\item[(iv)] Model learning.
Last, we leverage task-specific loss functions and a gradient descent procedure to train the model. 
\end{enumerate}

Designing a feasible hierarchical structure is crucial for HMGNNs, as the hierarchical structure determines how messages can be passed through different levels and what kind of meso- and macro-level information to be encoded in node representations.
In this paper, we consider (but are not restricted to) \textit{network communities}.
As a natural graph property, the community has been proved very useful for many graph mining tasks~\cite{WPL14,WCWP0Y17}.
Lots of community detection methods can generate hierarchical community structures.
Here, we propose an implementation model for the proposed framework, \textit{Hierarchical Community-aware Graph Neural Network} (HC-GNN).
HC-GNN exploits a well-known hierarchical community detection method, i.e., the \textit{Louvain} method~\cite{BGLL08} to build up the hierarchical structure, which is then used for the hierarchical message-passing mechanism.

The theoretical analysis illustrates HC-GNN's remarkable capacity in capturing long-range information without introducing heavy additional computation complexity.
Extensive empirical experiments are conducted on $9$ graph datasets to reveal the performance of HC-GNN on a variety of tasks, i.e., link prediction, node classification, and community detection, under transductive, inductive and few-shot settings. 
The results show that HC-GNN consistently outperforms a set of state-of-the-art approaches for link prediction and node classification.
In the few-shot learning setting, where only $5$ samples of each label are used to train the model,
HC-GNN achieves a significant performance improvement, up to $16.4\%$.
We also deliver a few empirical insights:  
(a) the lowest level contributes most to node representations;
(b) how to generate the hierarchical structure has a significant impact on the quality of node representations;
(c) HC-GNN maintains an outstanding performance for graphs with different levels of sparsity perturbation;
(d) HC-GNN possess significant flexibility in incorporating different GNN encoders, which means HC-GNN can achieve superior performance with advanced flat GNN encoders. 

\smallskip\noindent
\textbf{Contributions}.
The contribution of this paper is five-fold:
\begin{enumerate}
\item We propose a novel \textit{Hierarchical Message-passing Graph Neural Networks} framework,
which allows nodes to conveniently capture informative long-range interactions and encode multi-grained semantics hidden behind the given graph.
\item We present the first implementation of our framework, namely HC-GNN\footnote{Code and data are available at \url{https://github.com/zhiqiangzhongddu/HC-GNN}},
by detecting and utilising hierarchical community structures for message passing.
\item Theoretical analysis demonstrate the efficiency and the capacity of HC-GNN in capturing long-range interactions in graphs.
\item Experimental results show that HC-GNN significantly outperforms competing GNN methods on several prediction tasks under transductive, inductive, and few-shot settings.
\item Further empirical analysis is conducted to derive insights into the impact of the hierarchical structure and graph sparsity on HC-GNN and confirm its flexibility in incorporating different GNN encoders.
\end{enumerate}

The rest of this paper is organised as follows. 
We begin by briefly reviewing additional related work in Sec.~\ref{sec:related_work}.
Then in Sec.~\ref{sec:problem_statement}, we introduce the preliminaries of this study and state the research problem. 
In Sec.~\ref{sec:proposed_approach}, we introduce our proposed framework \textit{Hierarchical Message-passing Graph Neural Networks} and its first implementation, HC-GNN.
Experimental results and empirical analysis are shown in Sec.~\ref{sec:experiments}.
Finally, we conclude the paper and discuss the future work in Sec.~\ref{sec:conclusion_and_future_work}.


\section{Related Work}
\label{sec:related_work}
\noindent
\textbf{Flat message-passing GNNs.}
They perform graph convolution, directly aggregate node features from neighbours in the given graph,
and stack multiple GNN layers to capture long-range node dependencies~\cite{KW17,HYL17,VCCRLB18,XHLJ19}.
However, they were observed \textit{not} to benefit from more than a few layers, and recent studies have theoretically expressed this problem as \textit{over-smoothing}~\cite{LHW18,AY21}, i.e., node representations become indistinguishable when the number of GNN layers increases.
On the other hand, GraphRNA~\cite{HSLH19} presents graph recurrent networks to capture interactions between far-away nodes. 
Still, we cannot apply it to inductive learning settings because they rely on attributed random walks and the recurrent aggregations introduce high computation costs. 
P-GNNs~\cite{YYL19} incorporate a novel global information aggregation mechanism based on the distance of a given target node to each anchor set.
However, P-GNNs sacrifice the ability of existing GNNs on inductive node-wise tasks. 
As shown in their paper, they only support pairwise node classification tasks, i.e., comparing if two nodes have the same class label instead of predicting the class label of each individual node. 
Additionally, the anchor-set sampling operation brings a high computational cost for large-size graphs.
Recently, deeper \textit{flat} GNNs have been proposed by simplifying the aggregation design of GNNs and accompanied by well-designed normalisation units~\cite{CWHDL20} or specific gradient descent methods ~\cite{GCZSG20}.
Nevertheless, \cite{AY21} has theoretically shown that flat GNNs are susceptible to being a \textit{bottleneck} when aggregating messages across a long path and lead to severe \textit{over-squashing} issues. 
Moreover, we will theoretically discuss the advantages of our method compared with flat GNNs in Sec.~\ref{subsec:theoretical_analysis_and_model_comparison}, in terms of long-range interactive capability and complexity.

\smallskip\noindent
\textbf{Hierarchical representation GNNs.}
In recent years, some studies generalise the pooling mechanism of computer vision~\cite{RFB15} to GNNs for graph representation learning~\cite{YYMRHL18,HLLLL19,GJ19,RST20,LKALSBA20,LCZT20,RW21}.
However, most of them, such as \textsc{DiffPool}~\cite{YYMRHL18}, \textsc{AttPool}~\cite{HLLLL19} and ASAP~\cite{RST20}, are designed for graph classification tasks rather than learning node representations to capture long-range dependencies and multi-resolution semantics. 
Thus they cannot be directly applied to node-level tasks.
\textsc{g-U-Net}~\cite{GJ19} defines a similarity-based pooling operator to construct the hierarchical structure, and GXN~\cite{LCZT20} designs another infomax pooling operator, they implement bottom-up and top-down operations. 
Despite the success of \textsc{g-U-Net} and GXN in producing graph-level representations, they cannot model the multi-grained semantics and realise long-range message-passing.
HARP~\cite{CPHS18} and LouvainNE~\cite{BMDGM20} are two unsupervised network representation approaches that adopt a hierarchical structure, but they do not support the supervised training paradigm to optimise for specific tasks, and they cannot be applied with inductive settings.

More recently, HGNet~\cite{RW21} leverages multi-resolution representations of a graph to facilitate capturing long-range interactions.
Below, we discuss the main differences between HGNet and HC-GNN.
HC-GNN designs different efficient and effective bottom-up and top-down propagation mechanisms to realise elegant hierarchical message-passing rather than directly applying pooling and relational GCN, respectively. 
We further provide the theoretical analysis to demonstrate the efficiency and capacity of HC-GNN, such analysis has not been performed on HGNet. 
We also provide a much more careful and comprehensive set of experimental studies to validate the effectiveness of HC-GNN, including comparing learning settings on node classification (transductive, inductive, and few-sot), comparing to more recent competing flat GNN methods, comparing to state-of-the-art hierarchical GNN models, evaluating on the link prediction task, and in-depth analysis on graph sparsity and primary GNN encoders (Sec.~\ref{sec:experiments}).
Last but not least, in addition to capturing long-range interactions, we further deeply discuss the benefits and the usefulness of the hidden hierarchical structure in a graph.

Table~\ref{table:comparison_diff_GNNs} summarises the critical advantages of the proposed HC-GNN and compares it with a number of state-of-the-art methods published recently.
We are the first to present the hierarchical message passing to efficiently model long-range informative interaction and multi-grained semantics. 
In addition, our HC-GNN can utilise the community structures and be applied for transductive, inductive and few-shot inferences.


\section{Problem Statement}
\label{sec:problem_statement}
An attributed graph with $n$ nodes can be represented as $\mathcal{G}=(\mathcal{V}, \mathcal{E}, \mathbf{X})$, where $\mathcal{V}=\{v_{1}, v_{2},\dots, v_{n}\}$ is the node set, $\mathcal{E} \subseteq \mathcal{V} \times \mathcal{V}$ denotes the set of edges, and $\mathbf{X}=\{\mathbf{x}_{1}, \mathbf{x}_{2}, \dots, \mathbf{x}_{n}\}\in \mathbb{R}^{n \times \pi}$ is the feature matrix, in which each vector $\mathbf{x}_i\in \mathbf{X}$ is the feature vector associated with node $v_{i}$, and $\pi$ is the dimension of input feature vector of each node.
For subsequent discussion, we summarise $\mathcal{V}$ and $\mathcal{E}$ into an adjacency matrix $\mathbf{A} \in \{0, 1\}^{n \times n}$.

\smallskip\noindent
\textbf{Problem definition.}
Given a graph $\mathcal{G}$ and a pre-defined representation dimension $d$,
the goal is to learn a mapping function $f:\! \mathcal{G} \to \mathbf{Z}$, where $\mathbf{Z} \in \mathbb{R}^{n \times d}$ and each row $\mathbf{z}_{i}\in \mathbf{Z}$ corresponds to the node $v_i$'s representation.
The effectiveness of $f$ is evaluated by applying $\mathbf{Z}$ to different tasks, including node classification, link prediction, and community detection.
Table~\ref{table:summary_notations} lists the mathematical notation used in the paper.

\begin{table}[!ht]
\caption{Summary of main notations.
}
\label{table:summary_notations}
\centering
\begin{tabular}{l | l }
\hline
Notation & Description   \\
\hline
$\mathcal{G}$ & an attributed graph   \\   
$\mathcal{V}, \mathcal{E}$ & the set of nodes and edges on $\mathcal{G}$, respectively  \\	
$\mathbf{A}$ & the adjacent matrix of $\mathcal{G}$  \\	
$\mathbf{X} \in \mathbb{R}^{n \times \pi}$ & the matrix of node features   \\
$d$ & the pre-defined representation dimension   \\
$\mathbf{H} \in \mathbb{R}^{n \times d}$ & the hidden node representation matrix     \\
$\mathbf{h}_v \in \mathbb{R}^{d}$ & the hidden node representation of node $v$     \\
$\mathbf{Z} \in \mathbb{R}^{n \times d}$ & the final node representation matrix     \\
$\mathbf{z}_v \in \mathbb{R}^{d}$ & the final node representation of node $v$     \\
$L$ & the number of layers of \textit{within-level propagation} GNN encoder   \\
$T$ & the number of hierarchy levels     \\
$\mathcal{G}_{t}$ & the super graph at level $t$   \\
$s^{t}_{n}$ & the $n$-th super node of $\mathcal{G}_{t}$ at level $t$   \\
$\mathcal{H}$ & the set of constructed super graphs   \\
$\mathcal{N}(v)$ & the set of neighbour nodes of node $v$    \\
$\gamma$ & a hyper-parameter that used to construct super graph $\mathcal{G}_t$     \\
$\lambda$ & the pooling ratio     \\
\hline
\end{tabular}
\vspace{-5mm}
\end{table}

\smallskip\noindent
\textbf{Flat node representation learning}.
Prior to introducing the hierarchical message-passing mechanism, we first give a general review of existing Graph Neural Networks (GNNs) with \textit{flat} message-passing.
Let $\hat{\mathbf{A}} = (\hat{\mathbf{A}}_{uv})_{u,v \in \mathcal{V}}$, where $\hat{\mathbf{A}}_{uv}$ is a normalised value of $\mathbf{A}_{uv}$. 
Thus, we can formally define $\ell$-th layer of a flat GNN as:
\begin{equation}
\label{eq:gnn_aggregate_combine}
\begin{aligned}
\mathbf{m}^{(\ell)}_a &= \textsc{Aggregate}^{N}(\{\hat{\mathbf{A}}_{uv}, \, \mathbf{h}^{(\ell-1)}_u \, \vert \, u \in \mathcal{N}(v) \}), \\
\mathbf{m}^{(\ell)}_v &= \textsc{Aggregate}^{I}(\{\hat{\mathbf{A}}_{uv} \, \vert \, u \in \mathcal{N}(v) \}) \, \mathbf{h}^{(\ell-1)}_v, \\
\mathbf{h}^{(\ell)}_v &= \textsc{Combine}(\mathbf{m}^{(\ell)}_a, \mathbf{m}^{(\ell)}_v)
\end{aligned}
\end{equation}
where $\textsc{Aggregate}^{N}(\cdot)$ and $\textsc{Aggregate}^{I}(\cdot)$ are two possibly differential parameterised functions. 
$\mathbf{m}^{(\ell)}_a$ is aggregated message from node $v$'s neighbourhood nodes ($\mathcal{N}(v)$) with their structural coefficients, and $\mathbf{m}^{(\ell)}_v$ is the residual message from node $v$ after performing an adjustment operation to account for structural effects from its neighbourhood nodes.
After, $\mathbf{h}^{(\ell)}_v$ is the learned as representation vector of node $v$ by with combining $\mathbf{m}^{(\ell)}_a$ and $\mathbf{m}^{(\ell)}_v$, termed as $\textsc{Combine}(\cdot)$, at the $\ell$-th iteration/layer.
Note that, we initialise $\mathbf{h}^{(0)}_v = \mathbf{x}_v$ and the final learned representation vector after $L$ iterations/layers $\mathbf{z}_v = \mathbf{h}^{(L)}_v$.

Take the classic Graph Convolutional Network (GCN)~\cite{KW17} as an example, which applies two normalised mean aggregations to aggregate feature vectors node $v$'s neighbourhood nodes $\mathcal{N}(v)$ and combine with itself:
\begin{equation}
\label{eq:def_gcn}
    \mathbf{h}^{(\ell)}_v = \mathrm{ReLU}(\sum\limits_{u \in \mathcal{N}(v)}\frac{\mathbf{W}^{(\ell)} \mathbf{h}^{(\ell-1)}_u}{\sqrt{\vert \mathcal{N}(u) \vert \vert \mathcal{N}(v) \vert}} + \frac{\mathbf{W}^{(\ell)}\mathbf{h}^{(\ell-1)}_v}{\sqrt{\vert \mathcal{N}(v) \vert \vert \mathcal{N}(v) \vert}})
\end{equation}
where $\sqrt{\vert \mathcal{N}(u) \vert \vert \mathcal{N}(v) \vert}$ is a constant normalisation coefficient for the edge $\mathcal{E}_{uv}$, which is calculated from the normalised adjacent matrix $\mathbf{D}^{-1/2}\mathbf{A}\mathbf{D^{-1/2}}$. 
$\mathbf{D}$ is the diagonal node degree matrix of $\mathbf{A}$. 
$\mathbf{W}^{(\ell)} \in \mathbb{R}^{n \times d}$ is a trainable weight matrix of layer $\ell$.
From Eq.~\ref{eq:gnn_aggregate_combine} and Eq.~\ref{eq:def_gcn}, we can find that existing GNNs iteratively pass messages between adjacent nodes along observed edges, which will lead to two significant limitations: (a) the limited ability for information aggregation over long-range. They need to stack $k$ layers to capture interactions within $k$ steps for each node; (b) they are infeasible in encoding meso- and macro-level graph semantics.


\section{Proposed Approach}
\label{sec:proposed_approach}
\begin{figure*}[!ht]
\centering
\includegraphics[width=1.\linewidth]{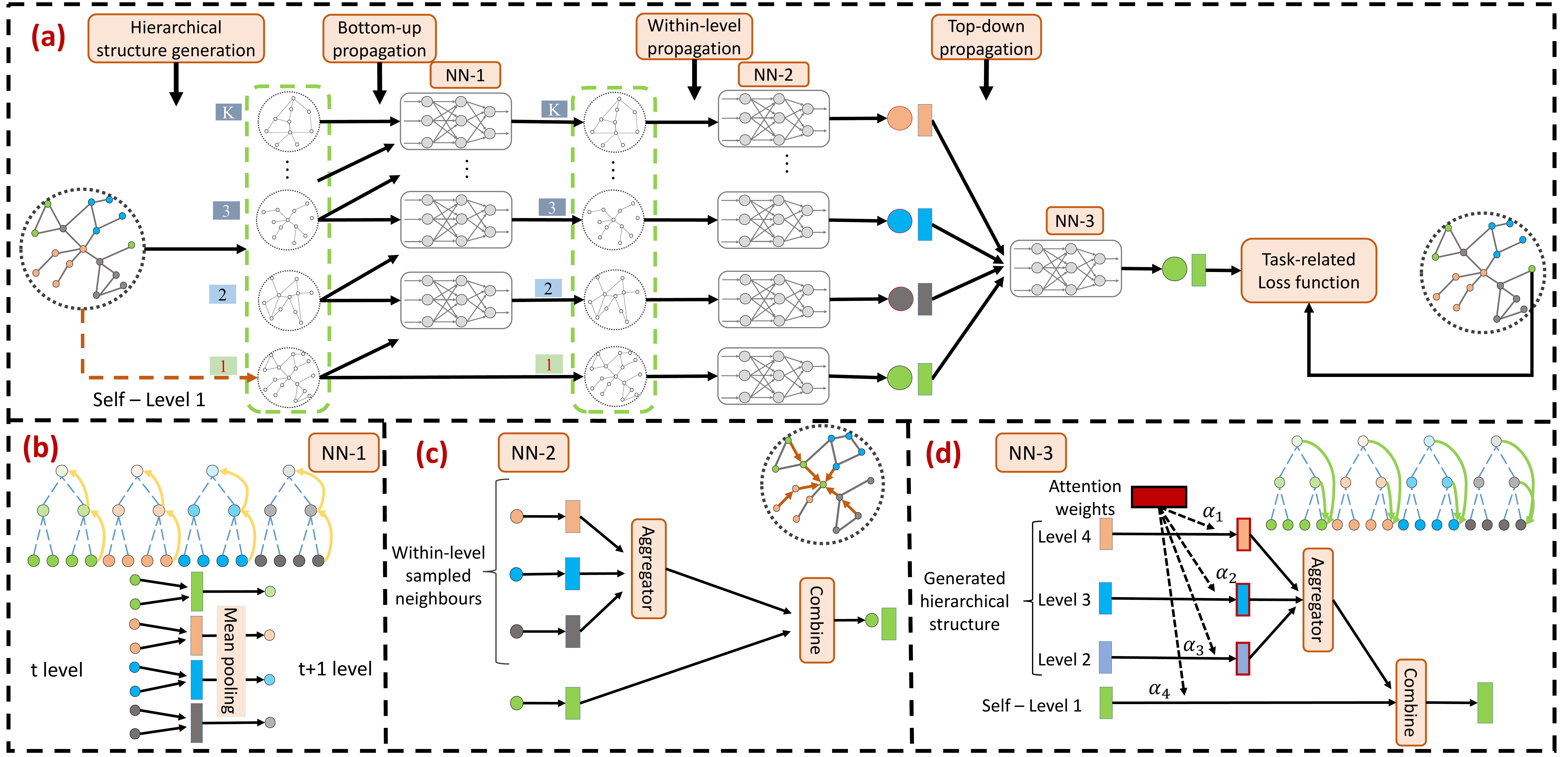}
\caption{
(a) The architecture of \textit{Hierarchical Message-passing Graph Neural Networks}: we first generate a hierarchical structure, in which each level is formed as a super graph, use the level $t$ graph to update nodes of level $t+1$ graph (bottom-up propagation), apply the typical neighbour aggregation on each level's graph (within-level propagation), use the generated node representations from level $2\leq t\leq T$ to update node representations at the level $1$ (top-down propagation), and optimises the model via a task-specific loss.
(b) NN-1: bottom-up propagation.
(c) NN-2: within-level propagation.
(d) NN-3: top-down propagation.
}
\label{fig:architesture_H-GNNs}
\vspace{-3mm}
\end{figure*}

We propose a framework, \textit{Hierarchical Message-passing Graph Neural Networks} (HMGNNs), whose core idea is to use a hierarchical message-passing structure to enable node representations to receive long-range messages and multi-grained semantics from different levels.
Fig.~\ref{fig:architesture_H-GNNs} provides an overview of the proposed framework, consisting of four components.
First, we create a hierarchical structure to coarsen the input graph $\mathcal{G}$ gradually.
Nodes at each level $t$ of the hierarchy are grouped into different super nodes ($s^{t}_{1}, \dots, s^{t}_{n}$).
Second, we further organise level $t$ generated super nodes into a super graph $\mathcal{G}_{t+1}$ at level $t\!+\!1$ based on the connections between nodes at level $t$, in order to enable message-passing that encodes the interactions between generated super nodes.
Third, we develop three different propagation schemes to propagate messages among nodes within the same level and across different levels.
At last, after obtaining node representations, we use the task-specific loss function and a gradient descent procedure to train the model.

\subsection{Hierarchical Message-passing GNNs}
\label{subsec:hierarchical_message-passing_graph_neural_networks}

\smallskip\noindent
\textbf{I. Hierarchical structure generation.}
Nodes $\mathcal{V}$ of a graph $\mathcal{G}$ can be naturally organised by super node structures of $T$ different levels, i.e., $\{\mathcal{V}_1, \mathcal{V}_2, \dots, \mathcal{V}_T \}$, in which densely inter-connected nodes of $\mathcal{V}_{t\!-\!1}$ ($2\le t\le T$) are grouped into a super node of $\mathcal{V}_{t}$. 
For example in Fig.~\ref{fig:general_structure}-(a), author set $\mathcal{V}_1=\{v_{1}, v_{2}, \dots, v_{17}\}$ can be grouped into different super nodes $\mathcal{V}_2=\{s_{1}, s_{2}, \dots, s_{9}\}$ based on their institutes.
Institutes can be further grouped into higher-level super nodes $\mathcal{V}_3=\{r_{1}, r_{2}, \dots, r_{4}\}$ according to research areas.
Meanwhile, there is a relationship between nodes at different levels, as indicated by dashed lines in Fig.~\ref{fig:general_structure}-(c).
Hence, we can generate a hierarchical structure to depict the inter- and intra-relationships among authors, institutes, and research areas. 
We will discuss how to implement the hierarchical structure generation in Sec.~\ref{subsec:hierarchical_community_aware_graph_neural_network}.

\smallskip\noindent
\textbf{II. $t$-Level super graph construction.}
The level $t$'s super graph $\mathcal{G}_{t}$ is constructed based on level $t\!-\!1$ graph $\mathcal{G}_{t\!-\!1}$ ($t\ge 2$), where $\mathcal{G}_1$ represents the original graph $\mathcal{G}$.
Given nodes at level $t\!-\!1$, i.e., $\mathcal{V}_{t\!-\!1}=\{s^{t\!-\!1}_{1}, \dots, s^{t\!-\!1}_{m}\}$, 
densely inter-connected nodes of $\mathcal{V}_{t\!-\!1}$ are grouped into a super node of $\mathcal{V}_{t}$ according to Sec.~\ref{subsec:hierarchical_message-passing_graph_neural_networks}-I.
We further create an edge between two super nodes $s^{t}_{i}$ and $s^{t}_{j}$ if there exist more than $\gamma$ edges in $\mathcal{G}_{t\!-\!1}$ connecting elements in $s^{t}_{i}$ and elements in $s^{t}_{j}$, where $\gamma$ is a hyper-parameter and $\gamma=1$ by default.
In this way, we can have an alternative representation of the hierarchical structure as a list of (super) graphs $\mathcal{H}=\{\mathcal{G}_{1}, \dots, \mathcal{G}_{T}\}$, where $\mathcal{G}_{1} = \mathcal{G}$. 
Moreover, inter-level edges are created to depict the relationships between (super) nodes at different levels $t$ and $t\!-\!1$, if a level $t\!-\!1$ node has a corresponding super node at level $t$, see for example Fig.~\ref{fig:general_structure}-(c).
We initialise the feature vectors of generated super nodes to be zero vectors with the same length as the original node feature vector $\mathbf{x}_i$.
Taking the collaboration network in Fig.~\ref{fig:general_structure} as an example, at the micro-level (level 1), we have authors and their co-authorship relations; at the meso-level (level 2), we organise authors according to their affiliations and establish relations between institutes; at the macro-level (level 3), institutes are further grouped according to their research areas, and we have the relations among the research areas.
In addition, inter-level links are also created to depict the relationships between authors and institutes and between institutes and research areas.

\smallskip\noindent
\textbf{III. Hierarchical message propagation.}
The hierarchical message-passing mechanism works as a supplementary process to enhance the node representations with long-range interactions and multi-grained semantics. 
Thus it does not change the flat node representation learning process as described in Sec.~\ref{sec:problem_statement}, to ensure the local information is well maintained. 
And we adopt the classic GCN, as described in Eq.~\ref{eq:def_gcn}, as our default flat GNN encoder throughout the paper.
Particularly, the hierarchical message-passing mechanism consists of $\ell$-th layer consisting of $3$ steps.

\begin{enumerate}
\item
\textit{Bottom-up propagation.}
After obtaining node representations ($\mathbf{h}^{(\ell)}_{s^{t\!-\!1}}$) of $\mathcal{G}_{t\!-\!1}$ with $\ell$-th flat information aggregation,
we perform bottom-up propagation, i.e., NN-1 in Fig.~\ref{fig:architesture_H-GNNs}-(b), using node representations in $\mathcal{G}_{t\!-\!1}$ to update node representations in $\mathcal{G}_{t}$ ($t\ge 2$) in the hierarchy $\mathcal{H}$, as follows:
\begin{equation}
	\mathbf{a}^{(\ell)}_{s^{t}_i} = \frac{1}{\vert{s^{t}_i}\vert+1} \left(\sum_{s^{t\!-\!1} \in s^{t}_i} \mathbf{h}^{(\ell)}_{s^{t\!-\!1}} + \mathbf{h}^{(\ell\!-\!1)}_{s^{t}_i}\right)
\end{equation}
where $s^{t}_i$ is a super node in $\mathcal{G}_{t}$, and $s^{t\!-\!1}$ is a node in $\mathcal{G}_{t\!-\!1}$ that belongs to $s^{t}_i$ in $\mathcal{G}_{t}$.
$\mathbf{h}^{(\ell\!-\!1)}_{s^{t}_i}$ is the node representation of $s^{t}_i$ that generated by layer $\ell \!-\! 1$ in graph $\mathcal{G}_{t}$, $\vert {s^{t}_i} \vert$ is the number of nodes of level $t\!-\!1$ that belonging to super node $s^{t}_i$, 
and $\mathbf{a}^{(\ell)}_{s^{t}_i}$ is the updated representation of $s^{t}_i$.

\item
\textit{Within-level propagation.}
We explore the typical \textit{flat} GNN encoders~\cite{KW17,HYL17,VCCRLB18,XHLJ19,CWHDL20} to propagate information within each level's graph $\{\mathcal{G}_{1}, \mathcal{G}_{2},\dots, \mathcal{G}_{T} \}$, i.e., NN-2 in Fig.~\ref{fig:architesture_H-GNNs}-(c).
The aim is to aggregate neighbours' information and update within-level node representations.
Specifically, the information aggregation at level $t$ is depicted as follows:
\begin{equation}
\begin{aligned}
\mathbf{m}^{(\ell)}_a &= \textsc{Aggregate}^{N}(\{\hat{\mathbf{A}}^{t}_{uv}, \, \mathbf{a}^{(\ell)}_u \, \vert \, u \in \mathcal{N}^{t}(v) \}), \\
\mathbf{m}^{(\ell)}_v &= \textsc{Aggregate}^{I}(\{\hat{\mathbf{A}}^{t}_{uv} \, \vert \, u \in \mathcal{N}^{t}(v) \}) \, \mathbf{a}^{(\ell)}_{v}, \\
\mathbf{b}^{(\ell)}_v &= \textsc{Combine}(\mathbf{m}^{(\ell)}_a, \mathbf{m}^{(\ell)}_v) \\
\end{aligned}
\end{equation}
where $\mathbf{a}^{(\ell)}_{u}$ is the node representation of $u$ after bottom-up propagation at the $\ell$-th layer,
$\mathcal{N}^{t}(v)$ is a set of nodes adjacent to $v$ at level $t$, and $\mathbf{b}^{(\ell)}_{v}$ is the aggregated node representation of $v$ based on local neighbourhood information.
Note that we adopt the classic GCN, as described in Eq.~\ref{eq:def_gcn}, as our default GNN encoder throughout the paper. We will discuss the possibility of incorporating with other advanced GNN encoders in Sec.~\ref{subsec:empirical_model_analysis}.

\item
\textit{Top-down propagation.}
The top-down propagation is illustrated by NN-3 in Fig.~\ref{fig:architesture_H-GNNs}-(d). 
We use node representations in $\{\mathcal{G}_{2}, \dots, \mathcal{G}_{T}\}$ to update the representations of original nodes in $\mathcal{G}$.
The importance of messages at different levels can be different for other tasks. 
Hence, we adopt the attention mechanism~\cite{VCCRLB18} to adaptively learn the contribution weights of different levels during top-down integration, given by: 
\begin{equation}
	\mathbf{h}^{(\ell)}_{v} = \mathrm{ReLU}(\mathbf{W} \cdot \textsc{MEAN} \{ \alpha_{uv} \mathbf{b}^{(\ell)}_{u} \}), \forall u \in \mathcal{C}(v) \cup{\{v\}}
\end{equation}
where $\alpha_{uv}$ is a trainable normalised attention coefficient between node $v$ to super node $u$ or itself,
$\textsc{MEAN}$ is an element-wise mean operation,
$\mathcal{C}(v)$ denotes the set of different-level super nodes from level $\{2, \dots, K\}$ that node $v$ belongs to ($\vert \mathcal{C}(v) \vert =K-1$), and $\mathrm{ReLU}$ is the activation function. 
$\mathbf{H}^{(\ell)}$ is the generated node representation of layer $\ell$ with $\mathbf{h}^{(\ell)}_{v} \in \mathbf{H}^{(\ell)}$.
We generate the output node representations of the last layer ($L$) via: 
\begin{equation}
	\mathbf{z}_{v} = \sigma(\mathbf{W} \cdot \textsc{MEAN} \{ \alpha_{uv} \mathbf{b}^{(L)}_{u} \}), \forall u \in \mathcal{C}(v) \cup{\{v\}}
\end{equation}
where $\sigma$ is the Euclidean normalisation function to reshape values into $[0, 1]$.
$\mathbf{Z} \in \mathbb{R}^{n \times d}$ is the final generated node representation with each row vector $\mathbf{z}_{v} \in \mathbf{Z}$.
\end{enumerate}

\smallskip
\noindent
\textbf{IV. Model learning.}
The proposed HMGNNs could be trained in unsupervised, semi-supervised, or supervised settings.
Here, we only discuss the supervised setting used for node classification in our experiments.
We define the loss function based on cross entropy, as follows:
\begin{equation} \label{eq:loss_fun}
	\mathcal{L} = -\sum_{v\in \mathcal{V}} \mathbf{y}^{\top}_{v} \log (\mathrm{Softmax} (\mathbf{z}_{v}))
\end{equation}
where $\mathbf{y}_{v}$ is a one-hot vector denoting the label of node $v$.
We allow $\mathcal{L}$ to be customised for other task-specific objective functions, e.g., the negative log-likelihood loss~\cite{VCCRLB18}.

\begin{algorithm}[!ht]
\caption{
\textit{Hierarchical Message-passing Graph Neural Networks}
}
\label{alg:framework_hc_gnn}
\hspace*{\algorithmicindent} \textbf{Input:} Graph $\mathcal{G}=(\mathcal{V}, \mathcal{E}, \mathbf{X})$ \\
\hspace*{\algorithmicindent} \textbf{Output:} Node representations $\mathbf{Z} \in \mathbb{R}^{n \times d}$
\begin{algorithmic}[1]
\State $\mathbf{h}^{(0)}_{v} \gets \mathbf{x}_{v}$
\State Generate hierarchical structure: $\mathcal{H} = \{\mathcal{G}_{t} \; \vert \; t=1, 2, \dots, T\}$
\For{$\ell \gets \{1,2,\dots, L \}$}{
    \State $\mathbf{h}^{(\ell)}_v = \mathrm{ReLU}(\sum\limits_{u \in \mathcal{N}(v)}\frac{\mathbf{W}^{(\ell)} \mathbf{h}^{(\ell-1)}_u}{\sqrt{\vert \mathcal{N}(u) \vert \vert \mathcal{N}(v) \vert}} + \frac{\mathbf{W}^{(\ell)}\mathbf{h}^{(\ell-1)}_v}{\sqrt{\vert \mathcal{N}(v) \vert \vert \mathcal{N}(v) \vert}})$, $\forall v \in \mathcal{G}$
	\For{$t \gets \{2, \dots, T\}$}
		\State $\mathbf{a}^{(\ell)}_{s^{t}_i} = \frac{1}{\vert{s^{t}_i}\vert+1} \left(\sum_{s^{t\!-\!1} \in s^{t}_i} \mathbf{h}^{(\ell)}_{s^{t\!-\!1}} + \mathbf{h}^{(\ell\!-\!1)}_{s^{t}_i}\right)$, $\forall s^{t}_i \in \mathcal{G}_{t}$
		\State $\mathbf{b}^{(\ell)}_v = \mathrm{ReLU}(\sum\limits_{u \in \mathcal{N}(v)}\frac{\mathbf{W}^{(\ell)} \mathbf{a}^{(\ell)}_u}{\sqrt{\vert \mathcal{N}(u) \vert \vert \mathcal{N}(v) \vert}} + \frac{\mathbf{W}^{(\ell)}\mathbf{a}^{(\ell)}_v}{\sqrt{\vert \mathcal{N}(v) \vert \vert \mathcal{N}(v) \vert}})$, $\forall v \in \mathcal{G}_t$
	\EndFor
	\For{$v \in \mathcal{G}$}
	    \If{$\ell < L$}
	        \State $\mathbf{h}^{(\ell)}_{v} = \mathrm{ReLU}(\mathbf{W} \cdot \textsc{MEAN} \{ \alpha_{uv} \mathbf{b}^{(\ell)}_{u} \}), \forall u \in \mathcal{C}(v) \cup{\{v\}}$
	    \Else
	        \State $\mathbf{z}_{v} = \sigma(\mathbf{W} \cdot \textsc{MEAN} \{ \alpha_{uv} \mathbf{b}^{(L)}_{u} \}), \forall u \in \mathcal{C}(v) \cup{\{v\}}$
	    \EndIf
	\EndFor
}\EndFor
\end{algorithmic}
\end{algorithm}

We summarise the process of \textit{Hierarchical Message-passing Graph Neural Networks} in Algorithm~\ref{alg:framework_hc_gnn}.
Given a graph $\mathcal{G}$,
we first generate the hierarchical structure and combine it with the original graph $\mathcal{G}$, 
to obtain $\mathcal{H} = \{ \mathcal{G}_{t} \, \vert \, t=1, 2, \dots, T \}$, where $\mathcal{G}_{1}=\mathcal{G}$ (line $2$).
For each node, including original and generated super nodes, in each NN layer, we perform three primary operations in order:
(1) bottom-up propagation (line $6$),
(2) within-level propagation (line $7$), 
and (3) top-down propagation (line $9\!-\!15$).
After getting the representation vector of each node that is enhanced with informative long-range interactions and multi-grained semantics,
and we train the model with the loss function $\mathcal{L}$ in Eq.~\ref{eq:loss_fun}.

\subsection{Hierarchical Community-aware GNN}
\label{subsec:hierarchical_community_aware_graph_neural_network}
Identifying hierarchical super nodes for the proposed HMGNNs is the most crucial step as it determines how the information will be propagated within and between levels.
We consider \textit{hierarchical network communities} to construct the hierarchy. 
The network community has been proved helpful for assisting typical network analysis tasks, including node classification~\cite{WPL14,WCWP0Y17} and link prediction~\cite{SH12,RGPPG15}.
Taking the algorithm efficiency into account and avoiding introducing additional hyper-parameters, i.e., the number of hierarchy levels, we adopt the well-known \textit{Louvain} algorithm~\cite{BGLL08} to build the first implementation of HMGNNs, termed as \textit{Hierarchical Community-aware Graph Neural Network} (HC-GNN).
The \textit{Louvain} algorithm returns us a hierarchical structure as described in Sec.~\ref{subsec:hierarchical_message-passing_graph_neural_networks} without the need for a pre-defined number of hierarchies, based on which we can learn node representations involving long-range interactive information and multi-grained semantics. 
Due to page limit, we include more details about community detection algorithms in App.~\ref{sec:appendix_introduction_of_community_detection_algorithms}. 

\subsection{Theoretical Analysis and Model Comparison}
\label{subsec:theoretical_analysis_and_model_comparison}

\smallskip\noindent
\textbf{Long-range interactive capability}.
We now theoretically analyse the asymptotic complexity of different GNN models to capture long-range interaction. 
We first analyse flat GNN models, that they need to stack $\mathcal{O}(\mathrm{diam}(\mathcal{G}))$ layers to ensure the communication between any pair of nodes in $\mathcal{G}$. 
For HMGNNs, let us assume the pooling ratio $\lambda=\vert \mathcal{V}_{t\!+\!1}\vert / \vert \mathcal{V}_{t}\vert$. 
Thus, the potentially total number of nodes in HMGNNs over $\mathcal{G}$ with $n$ nodes is $\sum_{t=1}^{\infty} n \lambda^{t}=\mathcal{O}(n)$, while the number of possible levels is $\log_{\lambda^{-1}} n=\mathcal{O}(\log n)$. 
That said, the shortest path between any two nodes of $\mathcal{G}$ is upper-bounded by $\mathcal{O}(\log n)$. 
Compared to $\mathcal{O}(\mathrm{diam}(\mathcal{G}))$ with flat GNNs, HMGNNs leads to significant improvement over the capability in capturing long-range interactions. 

\smallskip\noindent
\textbf{Model complexity}.
For the vanilla flat GNN model, i.e., GCN, its computational complexity of one layer is $\mathcal{O}(n^{3})$~\cite{KW17}, and the computational complexity of a GCN model contains $\ell$ is $\mathcal{O}(\ell n^{3})$.
For another attention-enhanced flat GNN model, i.e., Graph Attention Network (GAT)~\cite{VCCRLB18}, except for the same convolutional operation as GCN, the additional masked attention over all nodes requires $\mathcal{O}(\ell n^{2})$ computational complexity~\cite{VCCRLB18}.
Thus, overall it takes $\mathcal{O}(\ell (n^{3} + n^{2}))$ complexity.
For the hierarchical representation model, graph U-Net (\textsc{g-U-Net})~\cite{GJ19}, its computational complexity of one hierarchy is $\mathcal{O}(2 \ell n^{3})$, because its unpooling operation introduces another $\mathcal{O}(\ell n^{3})$ complexity, in addition to the convolutional operations as GCN.
Thus the complexity of \textsc{g-U-Net} with $T$ levels is $\sum_{t=1}^{T}2\ell (n\lambda^{t\!-\!1})^{3} = \mathcal{O}(2\ell n^{3})$, since the pooled graphs are supposed have much smaller number of nodes than $\mathcal{G}$. 
For HC-GNN, take GCN as an example GNN encoder and the \textit{Louvain} algorithm as an example hierarchical structure construction method, which has optimal $O(n \log c)$ computational complexity~\cite{T15}, where $c$ is the average degree.
The top-down propagation allows each node of $\mathcal{G}$ to receive $T$ different messages from $T$ levels with different weights, this introduces $\mathcal{O}(Tn)$ computational complexity, where $T$ is the number of levels, and we assume $T \ll n$.
Altogether, the complexity of HC-GNN is $\sum_{t=1}^{T} \ell (n \lambda^{t\!-\!1})^{3} + \mathcal{O}(n \log c + Tn) = \mathcal{O}(\ell n^{3} + n \log c + Tn)$,
which is more efficient than GAT and \textsc{g-U-Net}.


\section{Experiments}
\label{sec:experiments}
We conduct extensive experiments to answer 6 research questions (RQ):
%
\begin{itemize}
\item \textbf{RQ1:} How does HC-GNN performs \textit{vs.} state-of-the-art methods for node classification (\textbf{RQ1-1}),  community detection \textbf{(RQ1-2)}, and link prediction (\textbf{RQ1-3})? 
\item \textbf{RQ2:} Can HC-GNN leads to satisfying performance under settings of transductive, inductive, and few-shot learning?
\item \textbf{RQ3:} How do different levels in the hierarchical structure contribute to the effectiveness of node representations?
\item \textbf{RQ4:} How do various hierarchical structure generation methods affect the performance of HC-GNN?
\item \textbf{RQ5:} Does HC-GNN survive from low sparsity of graphs? 
\item \textbf{RQ6:} Does HC-GNN available with different encoders? 
\end{itemize}

\subsection{Evaluation Setup}
\label{sub:datasets}

\begin{table}[!ht]
\caption{Summary of dataset statistics.
LP: Link Prediction,
NC: Node Classification,
CD: Community Detection,
N.A. means a dataset does not contain node features or node labels. 
}
\label{table:summary_datasets}
\centering
\begin{tabular}{c | c | c | c | c | c }
            \hline
Dataset     & Task      & \#Nodes   & \#Edges   & \#Features    & \#Classes \\
            \hline
Grid 	    & LP        & 400       & 760       & N.A.          & N.A.      \\   
Cora 	    & LP\&NC    & 2,708     & 5,278     & 1,433         & 7         \\	
Power       & LP        & 4,941     & 6,594     & N.A.          & N.A.      \\
Citeseer    & NC        & 3,312     & 4,660     & 3,703         & 6         \\
Pubmed      & NC        & 19,717    & 44,327    & 500           & 3         \\
Emails      & CD        & 799       & 10,182    & N.A.          & 18        \\
PPI         & NC        & 56,658    & 818,435   & 50            & 121       \\
Protein     & NC        & 42,576    & 79,482    & 29            & 3         \\
Ogbn-arxiv  & NC        & 169,343   & 1,166,243	& 128           & 40        \\
\hline
\end{tabular}
\end{table}

\smallskip\noindent
\textbf{Datasets.} We perform experiments on both synthetic and real-world datasets.
For the link prediction task, we adopt $3$ datasets: 
\begin{itemize}
    \item Grid~\cite{YYL19}. A synthetic $2$D grid graph representing a $20 \times 20$ grid with $\vert \mathcal{V} \vert =400$ and no node features.
    \item Cora~\cite{SNBGGE08}. A citation network consists of $2,708$ scientific publications and $5,429$ links.
    A $1,433$ dimensional word vector describes each publication as a node feature.
    \item Power~\cite{WS98}. An electrical grid of western US with $4,941$ nodes and $6,594$ edges and no node features.
\end{itemize}
For node classification, we use $6$ datasets: 
including Cora, Citeseer~\cite{KW17} and Pubmed~\cite{KW17} and a large-scale benchmark dataset Ogbn-arxiv~\cite{HFZDRLCL20} for transductive settings, 
and $2$ protein interaction networks Protein and PPI~\cite{YYMRHL18} for inductive settings.
\begin{itemize}
\item Cora. 
The same above-mentioned Cora dataset contains $7$ classes of nodes. 
Each node is labelled with the class it belongs to.
\item Citeseer~\cite{SNBGGE08}. 
Each node comes with $3,703$-dimensional node features. 
\item Pubmed~\cite{NLGH12}. A dataset consists of $19,717$ scientific publications from PubMed database about diabetes classified into one of $3$ classes.
Each node is described by a TF/IDF weighted word vector from a dictionary which consists of $500$ unique words. 
\item PPI~\cite{ZL17}. 
$24$ protein-protein interaction networks and nodes of each graph comes with $50$ dimensional feature vector. 
\item Protein~\cite{BOSVSK05}. 
$1113$ protein graphs and nodes of each graph comes with $29$ dimensional feature vector. 
Each node is labelled with a functional role of the protein.
\item Ogbn-arxiv~\cite{HFZDRLCL20}. 
A large-scale citation graph between $169,343$ computer science arXiv papers. Each node is an arXiv paper, and each directed edge indicates that one paper cites another one. 
Each paper comes with a $128$-dimensional feature vector obtained by averaging the embeddings of words in its title and abstract.
The task is to predict the $40$ subject areas of these papers. 
\end{itemize}
%
For node community detection, we use an email communication dataset: 
\begin{itemize}
    \item Emails~\cite{snapnets}. $7$ real-world email communication graphs from SNAP with no node features.
    Each graph has $6$ communities, and each node is labelled with the community it belongs to. 
\end{itemize}
The data statistics of datasets is summarised in Table~,\ref{table:summary_datasets} and they are available for download with our published code. 

\smallskip\noindent
\textbf{Experimental settings.}
We evaluate HC-GNN under the settings of transductive and inductive learning.
For node classification, we additionally conduct experiments with the few-shot setting. 

\begin{itemize}
\item
\textit{Transductive learning.}
For link prediction, we follow the experimental settings of~\cite{YYL19} to use $10\%$ existing links and an equal number of non-existent links as validation and test sets.
The remaining $80\%$ existing links and a dual number of non-existent links are used as the training set.
For node classification, we follow the semi-supervised settings of~\cite{KW17}: if there are enough nodes, for each class, we randomly sample $20$ nodes for training, $500$ nodes for validation, and $1000$ nodes for testing. 
For the Emails dataset, we follow the supervised learning settings of~\cite{HSLH19} to randomly select $80\%$ nodes as the training set, and use the two halves of remaining as the validation and test set, respectively. 
We report the test performance when the best validation performance is achieved.

\item
\textit{Inductive learning.}
This aims at examining a model's ability to transfer the learned knowledge from existing nodes to future ones that are newly connected to existing nodes in a graph.
Hence, we hide the validation and testing graphs during training.
We conduct the experiments for inductive learning using PPI and Protein datasets. 
We train models on $80\%$ graphs to learn an embedding function $f$ and apply it on the remaining $20\%$ graphs to generate the representation of new-coming nodes.

\item
\textit{Few-shot learning.}
Since the cost of collecting massive labelled datasets is high, having a few-shot learning model would be pretty valuable for practical applications. 
Few-shot learning can also be considered as an indicator to evaluate the robustness of a deep learning model.
We perform few-shot node classification, in which only $5$ samples of each class are used for training.
The sampling strategies for testing and validation sets follow those in transductive learning.
\end{itemize}

\smallskip\noindent
\textbf{Evaluation metrics.} 
We adopt AUC to measure the performance of link prediction.
For node classification, we use micro- and macro-average F1 scores and accuracy.
NMI score is utilised for community detection evaluation.

\smallskip\noindent 
\textbf{Competing methods.}
To validate the effectiveness of HC-GNN, we compare it with $10$ competing methods which include $6$ flat message-passing GNN models,
(GCN~\cite{KW17}, GraphSAGE~\cite{HYL17}, GAT~\cite{VCCRLB18}, GIN~\cite{XHLJ19}, P-GNNs~\cite{YYL19}, \textsc{GCNII}~\cite{CWHDL20}),
$3$ hierarchical GNN models 
(HARP~\cite{CPHS18}, \textsc{g-U-Net}~\cite{GJ19}, GXN~\cite{LCZT20}) 
and another state-of-the-art model.
(GraphRNA~\cite{HSLH19}).
For more details about competing methods, refer to App.~\ref{sec:appendix_competing_methods}. 

\smallskip\noindent 
\textbf{Reproducibility.}
For fair comparison, all methods adopt the same representation dimension ($d=32$), learning rate ($=1e\!-\!3$), Adam optimiser and the number of iterations ($=200$) with early stop ($50$). 
In terms of the neural network layers, we report the one with better performance of GCNII with better performance among $\{8, 16, 32, 64, 128\}$; for other models, we report the one with better performance between $2\!-\!4$;
For all models with hierarchical structure (including \textsc{g-U-Net} and HC-GNN), we use GCN as the default GNN encoder for fair comparision. 
Note that for the strong competitor, P-GNNs, since its representation dimension is related to the number of nodes in a graph, we add a linear regression layer at the end of P-GNNs for node classification tasks to ensure its end-to-end structure is the same as other models~\cite{HSLH19}.
For HC-GNN, the number of HC-GNN layers is varied and denoted as $1$L, $2$L or $3$L.
In Sec.~\ref{subsec:empirical_model_analysis}, HC-GNN adopts the number of layers leading to the best performance for model analysis i.e., $2$L for the Cora dataset, $1$L for the Citeseer and Pubmed datasets. 
For \textit{Louvain} community detection, we use the implementation of a given package\footnote{\url{https://python-louvain.readthedocs.io/en/latest/api.html}}, which does not require any hyper-parameters.
We use PyTorch Geometric to implement all models mentioned in this paper. 
More details are referred to our code file\footnote{Code and data are available at \url{https://github.com/zhiqiangzhongddu/HC-GNN}}.
The experiments are repeated $10$ times, and average results are reported.
Note that we use only node features with unique one-hot identifiers to differentiate different nodes if there are no given node features from the datasets and use the original node features if they are available.
We employ Pytorch to implement all models.
Experiments were conducted with GPU (NVIDIA Tesla V100) machines.

\subsection{Experimental Results}
\label{subsec:applications}

\begin{table*}[!ht]
\caption{Results in Micro-F1 and Macro-F1 for transductive semi-supervised node classification task. Results in Acc for node classification of Ogbn-arxiv follows the default settings of OGB dataset~\cite{HFZDRLCL20}, and results in NMI for community detection (i.e., on the Emails data in the last column). 
Standard deviation errors are given.
\textsuperscript{\ddag} indicates the results from OGB leaderboard~\cite{HFZDRLCL20}.
OOM: out-of-memory.
$1L$: model with $1$-layer GNN encoder for \textit{within-level propagation}.
}
\label{table:nc_results}
\centering
\resizebox{1.1\textwidth}{!}{
\begin{tabular}{l | c | c | c | c | c | c | c | c}
\hline
 			& \multicolumn{2}{c|}{\textbf{Cora}}    & \multicolumn{2}{c|}{\textbf{Citeseer}}    & \multicolumn{2}{c|}{\textbf{Pubmed}}  & \textbf{Emails} & \textbf{Ogbn-arxiv} \\
 			& Micro-F1          & Macro-F1          & Micro-F1          & Macro-F1              & Micro-F1          & Macro-F1          & NMI & Acc (\%) \\
\hline
\hline
GCN			& $0.802\scriptstyle\pm0.019$   & $0.786\scriptstyle\pm0.020$ 	& $0.648\scriptstyle\pm0.019$   & $0.612\scriptstyle\pm0.012$ 		& $0.779\scriptstyle\pm0.027$   & $0.777\scriptstyle\pm0.026$	& $0.944\scriptstyle\pm0.010$ & $71.74\scriptstyle\pm0.29^{\ddag}$	\\
GraphSAGE 	& $0.805\scriptstyle\pm0.013$ 	& $0.792\scriptstyle\pm0.009$ 	& $0.650\scriptstyle\pm0.027$   & $0.611\scriptstyle\pm0.020$		& $0.768\scriptstyle\pm0.031$	& $0.763\scriptstyle\pm0.030$	& $0.925\scriptstyle\pm0.014$ & $71.49\scriptstyle\pm0.27^{\dag}$	\\
GAT 		& $0.772\scriptstyle\pm0.019$ 	& $0.761\scriptstyle\pm0.023$ 	& $0.620\scriptstyle\pm0.024$   & $0.594\scriptstyle\pm0.015$	    & $0.775\scriptstyle\pm0.036$	& $0.770\scriptstyle\pm0.022$ 	& $\underline{0.947}\scriptstyle\pm0.009$ & $72.06\scriptstyle\pm0.31^{\dag}$	\\
GIN 		& $0.762\scriptstyle\pm0.020$ 	& $0.759\scriptstyle\pm0.018$	& $0.615\scriptstyle\pm0.023$   & $0.591\scriptstyle\pm0.020$ 		& $0.744\scriptstyle\pm0.036$	& $0.733\scriptstyle\pm0.041$ 	& $0.640\scriptstyle\pm0.047$ & $71.76\scriptstyle\pm0.33^{\dag}$ 	\\
P-GNNs 		& $0.438\scriptstyle\pm0.044$   & $0.431\scriptstyle\pm0.040$ 	& $0.331\scriptstyle\pm0.019$   & $0.314\scriptstyle\pm0.018$ 		& $0.558\scriptstyle\pm0.033$   & $0.551\scriptstyle\pm0.036$   & $0.598\scriptstyle\pm0.020$ & OOM	\\
GCNII 		& $\underline{0.823}\scriptstyle\pm0.017$   & $\underline{0.801}\scriptstyle\pm0.022$ 	& $\underline{0.722}\scriptstyle\pm0.011$   & $\underline{0.677}\scriptstyle\pm0.010$ 		& $\underline{0.791}\scriptstyle\pm0.009$   & $\underline{0.790}\scriptstyle\pm0.016$   & $0.947\scriptstyle\pm0.010$ & $\underline{72.74}\scriptstyle\pm0.16$	\\
HARP 	    & $0.363\scriptstyle\pm0.020$ 	& $0.350\scriptstyle\pm0.021$ 	& $0.343\scriptstyle\pm0.023$   & $0.317\scriptstyle\pm0.017$ 		& $0.441\scriptstyle\pm0.024$	& $0.329\scriptstyle\pm0.019$ 	& $0.371\scriptstyle\pm0.014$ & OOM	\\
GraphRNA 	& $0.354\scriptstyle\pm0.070$ 	& $0.244\scriptstyle\pm0.040$ 	& $0.352\scriptstyle\pm0.050$   & $0.259\scriptstyle\pm0.047$ 		& $0.476\scriptstyle\pm0.054$	& $0.355\scriptstyle\pm0.089$ 	& $0.434\scriptstyle\pm0.047$ & OOM	\\
\textsc{g-U-Net}    & $0.805\scriptstyle\pm0.017$ 	& $0.796\scriptstyle\pm0.018$ 	& $0.673\scriptstyle\pm0.015$   & $0.628\scriptstyle\pm0.012$ 		& $0.782\scriptstyle\pm0.018$	& $0.781\scriptstyle\pm0.019$ 	& $0.939\scriptstyle\pm0.015$ & $71.78\scriptstyle\pm0.37$ \\
GXN    & $0.811\scriptstyle\pm0.025$ 	& $\underline{0.801}\scriptstyle\pm0.031$ 	& $0.698\scriptstyle\pm0.017$   & $0.619\scriptstyle\pm0.021$ 		& $0.784\scriptstyle\pm0.014$	& $0.782\scriptstyle\pm0.012$ 	& $0.943\scriptstyle\pm0.011$ & $70.99\scriptstyle\pm0.27$ \\
\hline
HC-GNN-1$L$	& $0.819\scriptstyle\pm0.002$ 	& $\textbf{0.816}\scriptstyle\pm0.005$ 	& $\textbf{0.728}\scriptstyle\pm0.005$   & $\textbf{0.686}\scriptstyle\pm0.003$ 		& $\textbf{0.812}\scriptstyle\pm0.009$	& $\textbf{0.806}\scriptstyle\pm0.009$ 	& $\textbf{0.961}\scriptstyle\pm0.005$ & $72.69\scriptstyle\pm0.25$	\\
HC-GNN-2$L$	& $\textbf{0.834}\scriptstyle\pm0.007$ 	& $\textbf{0.816}\scriptstyle\pm0.006$ 	& $0.696\scriptstyle\pm0.002$   & $0.652\scriptstyle\pm0.006$ 		& $\textbf{0.809}\scriptstyle\pm0.004$	& $\textbf{0.804}\scriptstyle\pm0.005$	& $\textbf{0.962}\scriptstyle\pm0.005$ & $\textbf{72.79}\scriptstyle\pm0.31$ 	\\
HC-GNN-3$L$	& $0.813\scriptstyle\pm0.008$ 	& $\textbf{0.806}\scriptstyle\pm0.006$ 	& $0.686\scriptstyle\pm0.006$   & $0.633\scriptstyle\pm0.008$ 	    & $\textbf{0.804}\scriptstyle\pm0.004$	& $0.780\scriptstyle\pm0.020$ 	& $0.935\scriptstyle\pm0.014$ & $72.58\scriptstyle\pm0.27$	\\
\hline
\end{tabular}
}
\end{table*}

\smallskip\noindent
\textbf{Transductive node classification (RQ1-1\&RQ2).} 
We present the results of transductive node classification in Table~\ref{table:nc_results}.
We can see that HC-GNN consistently outperforms all of the competing methods in the $5$ datasets, and even the shallow HC-GNN model with only one layer may lead to better results.
We think the outstanding performance of HC-GNN results from two aspects: (a) the hierarchical structure allows the model to capture informative long-range interactions of graphs, i.e., propagating messages from and to distant nodes in the graph; and (b) the meso- and macro-level semantics reflected by the hierarchy is encoded through bottom-up, within-level, and top-down propagations.
On the other hand, P-GNNs, HARP, and GraphRNA perform worse in semi-supervised node classification.
The possible reason is they need more training samples, such as using $80\%$ of existing nodes as the training set, as described in their papers~\cite{YYL19,HSLH19}, but we have only $20$ nodes for training in the semi-supervised setting.

\begin{table}[!ht]
\caption{Micro-F1 results for inductive node classification. 
Standard deviation errors are given.
$1L$: model with $1$-layer GNN encoder for \textit{within-level propagation}.
}
\label{table:inductive_nc_results}
\centering
\begin{tabular}{l | c | c}
\hline
 			& \textbf{PPI}                  & \textbf{Protein}              \\
\hline
\hline
GCN			& $0.444\scriptstyle\pm0.004$               & $0.542\scriptstyle\pm0.018$               \\
GraphSAGE 	& $0.409\scriptstyle\pm0.014$			    & $\underline{0.637}\scriptstyle\pm0.018$   \\
GAT 		& $0.469\scriptstyle\pm0.062$ 			    & $0.608\scriptstyle\pm0.077$               \\
GIN 		& $\underline{0.571}\scriptstyle\pm0.008$ 	& $0.631\scriptstyle\pm0.016$               \\
GCNII 		& $0.507\scriptstyle\pm0.008$ 	& $0.614\scriptstyle\pm0.011$               \\
\textsc{g-U-Net}    & $0.433\scriptstyle\pm0.012$       & $0.547\scriptstyle\pm0.011$               \\
GXN         & $0.510\scriptstyle\pm0.094$               & $0.578\scriptstyle\pm0.014$               \\
\hline
HC-GNN-1$L$	& $0.48\scriptstyle\pm0.091$ 			    & $\textbf{0.638}\scriptstyle\pm0.027$      \\
HC-GNN-2$L$	& $\textbf{0.584}\scriptstyle\pm0.087$ 	    & $0.622\scriptstyle\pm0.031$               \\
HC-GNN-3$L$	& $\textbf{0.584}\scriptstyle\pm0.002$ 	    & $0.582\scriptstyle\pm0.025$               \\
\hline
\end{tabular}
\vspace{-3mm}
\end{table}

\smallskip\noindent
\textbf{Inductive node classification (RQ1-1\&RQ2)}.
The results are reported in Table~\ref{table:inductive_nc_results}\footnote{Since HARP, P-GNNs and GraphRNA cannot be applied in the inductive setting, we do not present their results in Table~\ref{table:inductive_nc_results}.}.
We can find that HC-GNN is still able to show some performance improvement over existing GNN models. 
But the improvement gain is not so significant and inconsistent in different layers of HC-GNN compared to the results in transductive learning. 
The possible reason is that different graphs may have other hierarchical community structures. 
Nevertheless, the results lead to one observation: the effect of transferring hierarchical semantics between graphs for inductive node classification is somewhat limited.
Therefore, exploring an ameliorated model that can adaptively exploit hierarchical structure for different graphs for different tasks would be interesting.
We further discuss it in Sec.~\ref{sec:conclusion_and_future_work} as one concluding remark. 

\begin{table}[!ht]
\caption{
Micro-F1 results for few-shot node classification. 
Standard deviation errors are given.
$1L$: model with $1$-layer GNN encoder for \textit{within-level propagation}.
}
\label{table:fs_nc_results}
\centering
\begin{tabular}{l | c | c | c}
\hline
 			& \textbf{Cora}                 & \textbf{Citeseer}             & \textbf{Pubmed}    \\
\hline
\hline
GCN			& $0.695\scriptstyle\pm0.049$ 	            & $0.561\scriptstyle\pm0.054$               & $0.699\scriptstyle\pm0.059$	                \\
GraphSAGE 	& $0.719\scriptstyle\pm0.024$ 	            & $0.559\scriptstyle\pm0.049$               & $0.707\scriptstyle\pm0.051$	                \\
GAT 		& $0.630\scriptstyle\pm0.030$ 	            & $0.520\scriptstyle\pm0.054$               & $0.664\scriptstyle\pm0.046$	                \\
GIN 		& $0.691\scriptstyle\pm0.038$ 	            & $0.509\scriptstyle\pm0.060$               & $0.714\scriptstyle\pm0.036$     \\
P-GNNs 		& $0.316\scriptstyle\pm0.040$ 	            & $0.332\scriptstyle\pm0.011$               & $0.547\scriptstyle\pm0.037$	                \\
GCNII 		& $0.701\scriptstyle\pm0.022$ 	            & $0.564\scriptstyle\pm0.015$               & $\underline{0.717}\scriptstyle\pm0.047$	                \\
HARP 	    & $0.224\scriptstyle\pm0.033$ 	            & $0.260\scriptstyle\pm0.035$               & $0.415\scriptstyle\pm0.039$ 	            \\
GraphRNA 	& $0.274\scriptstyle\pm0.063$ 	            & $0.206\scriptstyle\pm0.019$               & $0.429\scriptstyle\pm0.042$ 	            \\
\textsc{g-U-Net}    & $0.706\scriptstyle\pm0.054$ 	    & $\underline{0.567}\scriptstyle\pm0.044$   & $0.693\scriptstyle\pm0.036$	                \\
GXN         & $\underline{0.721}\scriptstyle\pm0.035$ 	& $0.564\scriptstyle\pm0.21$                & $0.706\scriptstyle\pm0.043$	                \\
\hline
HC-GNN-1$L$	& $0.681\scriptstyle\pm0.023$ 	            & $\textbf{0.639}\scriptstyle\pm0.019$      & $0.704\scriptstyle\pm0.043$	                \\
HC-GNN-2$L$	& $\textbf{0.759}\scriptstyle\pm0.015$ 	    & $\textbf{0.660}\scriptstyle\pm0.024$      & $\textbf{0.724}\scriptstyle\pm0.052$	    \\
HC-GNN-3$L$	& $\textbf{0.752}\scriptstyle\pm0.017$ 	    & $\textbf{0.642}\scriptstyle\pm0.016$      & $\textbf{0.742}\scriptstyle\pm0.045$	    \\
\hline
\end{tabular}
\vspace{-3mm}
\end{table}

\smallskip\noindent
\textbf{Few-shot node classification (RQ1-1\&RQ2)}.
Table~\ref{table:fs_nc_results} demonstrates better performance in few-shot learning than all competing methods across $3$ datasets. 
Such results indicate that the hierarchical message passing is able to transfer supervised information through inter- and intra-level propagations. 
In addition, the hierarchical message-passing pipeline further enlarges the influence range of supervision information from a small number of training samples. 
With effective and efficient pathways to broadcast information, HC-GNN is proven to be quite promising in few-shot learning.

\smallskip\noindent
\textbf{Community detection (RQ1-2).}
The community detection results conducted on the Emails dataset are also shown in Table~\ref{table:nc_results}. 
It can be seen that HC-GNN again outperforms all competing methods. 
We believe this is because the communities identified by Louvain are further exploited by learning their hierarchical interactions in HC-GNN. 
In other words, HC-GNN is able to reinforce the intra- and inter-community effect and encode it into node representations.

\begin{table}[!ht]
\caption{Results in AUC for link prediction.
Standard deviation errors are given.
$1L$: model with $1$-layer GNN encoder for \textit{within-level propagation}.
}
\label{table:lp_results}
\centering
\begin{tabular}{l | c | c | c | c | c}
\hline
 			& \textbf{Grid} 			    & \textbf{Cora-Feat}		    & \textbf{Cora-NoFeat}           & \textbf{Power-Feat} & \textbf{Power-NoFeat}  		\\
\hline
\hline
GCN			& $0.763\scriptstyle\pm0.036$ 				& $0.869\scriptstyle\pm0.006$ 				& $0.785\scriptstyle\pm0.007$               & $0.624\scriptstyle\pm0.013$ 				    & $0.562\scriptstyle\pm0.012$ \\
GraphSAGE 	& $0.775\scriptstyle\pm0.018$				& $0.870\scriptstyle\pm0.006$ 	            & $0.741\scriptstyle\pm0.017$               & $0.569\scriptstyle\pm0.012$ 				    & $0.510\scriptstyle\pm0.009$ \\
GAT 		& $0.782\scriptstyle\pm0.028$ 				& $0.874\scriptstyle\pm0.010$ 				& $0.789\scriptstyle\pm0.012$               & $0.621\scriptstyle\pm0.013$ 				    & $0.551\scriptstyle\pm0.019$ \\
GIN 		& $0.756\scriptstyle\pm0.025$ 				& $0.862\scriptstyle\pm0.009$ 				& $0.782\scriptstyle\pm0.010$               & $0.620\scriptstyle\pm0.011$ 				    & $0.549\scriptstyle\pm0.006$ \\
P-GNNs 		& $\underline{0.867}\scriptstyle\pm0.034$ 	& $0.818\scriptstyle\pm0.013$ 				& $\underline{0.792}\scriptstyle\pm0.012$   & $\underline{0.704}\scriptstyle\pm0.006$ 	    & $\underline{0.668}\scriptstyle\pm0.021$ \\
GCNII 		& $0.807\scriptstyle\pm0.024$               & $0.889\scriptstyle\pm0.019$ 				& $0.770\scriptstyle\pm0.011$               & $0.695\scriptstyle\pm0.014$ 	                & $0.577\scriptstyle\pm0.015$ \\
HARP 		& $0.687\scriptstyle\pm0.021$               & $0.837\scriptstyle\pm0.033$ 				& $0.721\scriptstyle\pm0.017$               & $0.529\scriptstyle\pm0.004$ 	                & $0.502\scriptstyle\pm0.004$ \\
\textsc{g-U-Net}    & $0.701\scriptstyle\pm0.032$               & $\underline{0.909}\scriptstyle\pm0.006$   & $0.772\scriptstyle\pm0.007$               & $0.628\scriptstyle\pm0.024$                     & $0.584\scriptstyle\pm0.019$ \\
GXN    & $0.642\scriptstyle\pm0.089$               & $0.889\scriptstyle\pm0.003$        & $0.781\scriptstyle\pm0.011$               & $0.645\scriptstyle\pm0.013$                     & $0.562\scriptstyle\pm0.015$ \\
\hline
HC-GNN-1$L$ 	& $0.823\scriptstyle\pm0.035$ 				& $0.884\scriptstyle\pm0.006$ 		        & $\textbf{0.795}\scriptstyle\pm0.012$      & $0.682\scriptstyle\pm0.016$        		        & $0.654\scriptstyle\pm0.017$ \\
HC-GNN-2$L$ 	& $\textbf{0.913}\scriptstyle\pm0.011$ 		& $0.895\scriptstyle\pm0.007$ 		        & $\textbf{0.837}\scriptstyle\pm0.006$      & $\textbf{0.767}\scriptstyle\pm0.020$ 		    & $\textbf{0.722}\scriptstyle\pm0.020$ \\
HC-GNN-3$L$ 	& $\textbf{0.914}\scriptstyle\pm0.011$ 		& $0.891\scriptstyle\pm0.007$ 		        & $\textbf{0.839}\scriptstyle\pm0.004$      & $\textbf{0.784}\scriptstyle\pm0.017$ 		    & $\textbf{0.746}\scriptstyle\pm0.021$ \\
\hline
\end{tabular}
\vspace{-3mm}
\end{table}

\smallskip\noindent
\textbf{Link prediction (RQ1-3).}
Here, we motivate our idea by considering pairwise relation prediction between nodes.
Suppose a pair of nodes $u,v$ are labelled with label $y$, and our goal is to predict $y$ for unseen pairs.
From the perspective of representation learning, we can solve the problem via learning an embedding function $f$ that computes the node representation $\mathbf{z}_{v}$, where the objective is to maximise the likelihood of distribution $p(y \vert \mathbf{z}_{u}, \mathbf{z}_{v})$.
The results in Table~\ref{table:lp_results} indicate that the HC-GNN leads to competitive performance compared to all competing methods, with up to $11.7\%$ AUC improvement, demonstrating its effectiveness on link prediction tasks. 
%
%
%
%
When node features are accessible (i.e., Cora-Feat and Power-Feat), all models perform relatively well, and \textsc{g-U-Net} has slightly better performance on Cora-Feat dataset. 
Because node features provide meaningful information to predict pairwise relations. 
Another interesting perspective is investigating the models' performance without contextual node features (e.g., Grid, Cora-NoFeat and Power-NoFeat). 
It is surprising that HC-GNN variants show great superiority in these three datasets.
We argue that when only topological information is available, the hierarchical semantics introduced by HC-GNN helps find missing links.

\subsection{Empirical Model Analysis}
\label{subsec:empirical_model_analysis}

\begin{figure}[!ht]
\centering
\includegraphics[width=1.\linewidth]{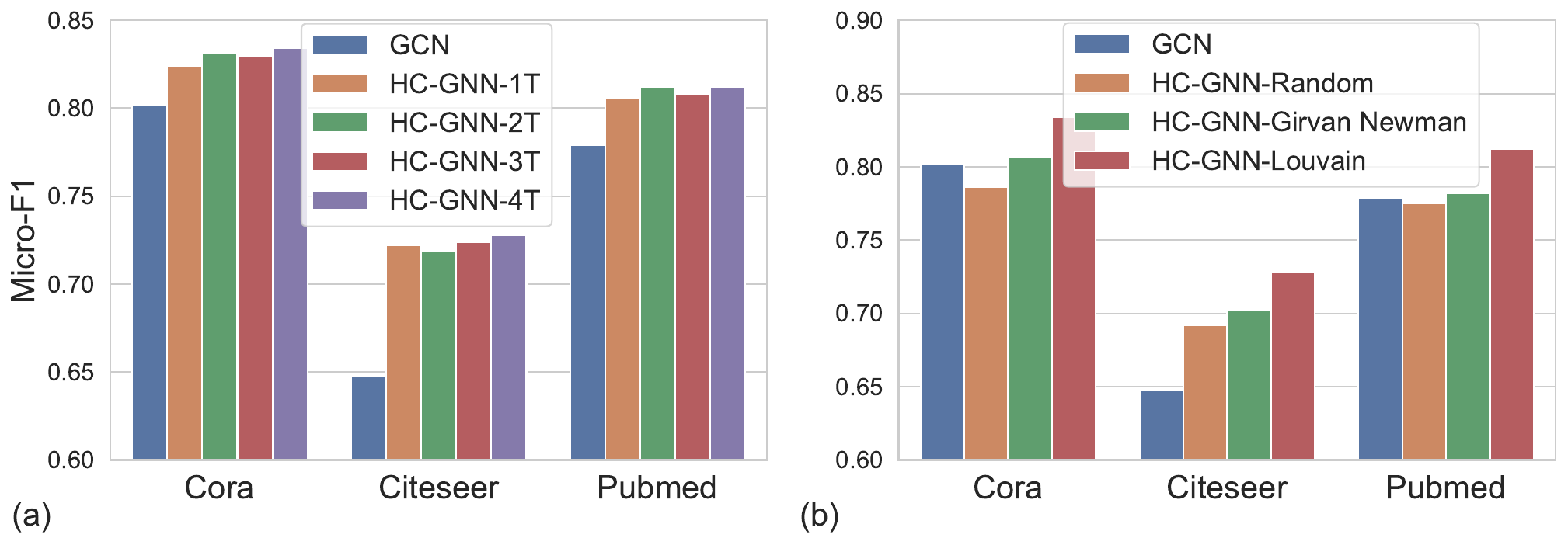}
\caption{Results in Micro-F1 for semi-supervised node classification using HC-GNN
by varying:
(a) the number of hierarchy levels adopted for message passing, and
(b) the approaches to generate the hierarchical structure.
$2T$ means model with first $2$ hierarchy levels. 
}
\label{fig:influence_comparisons}
\vspace{-3mm}
\end{figure}

\smallskip\noindent
\textbf{Contribution of different levels (RQ3).}
Since HC-GNN highly relies on the generated hierarchical structure, we aim to examine how different levels in the hierarchy contribute to the prediction.
We report the transductive semi-supervised node classification performance by varying the number of levels (from $1T$ to $4T$).
GCN is also selected for comparison because it considers no hierarchy, i.e., only within-level propagation in the original graph.
The results are shown in Fig.~\ref{fig:influence_comparisons}(a), in which $1T$ and $2T$ indicate only the first hierarchy level and the first $2$ hierarchy levels are adopted, respectively. 
We can find that HC-GNN using more levels for hierarchy construction lead to better results.
The flat message passing of GCN cannot work well.
Such results provide strong evidence that GNNs can significantly benefit from the hierarchical message-passing mechanism. 
In addition, more hierarchical semantics can be encoded if more levels are adopted.

\smallskip\noindent
\textbf{Influence of hierarchy generation approaches (RQ4)}.
HC-GNN implements the proposed \textit{Hierarchical Message-passing Graph Neural Networks} based on the \textit{Louvain} community detection algorithm, that is termed HC-GNN-\textit{Louvain} in this paragraph. 
We aim to validate (A) whether the community information truly benefits the classification tasks, and (B) how different approaches to generate the hierarchical structure affect the performance.
To answer (A), we construct a random hierarchical structure to generate randomised HC-GNN, termed HC-GNN-Random, in which \textit{Louvain} detects hierarchical communities, and nodes are randomly swapped among the same-level communities. 
In other words, the hierarchy structure is maintained, but community memberships are perturbed. 
The results on semi-supervised node classification are exhibited in Fig.~\ref{fig:influence_comparisons}(b). 
We can see that HC-GNN-Random works worse than GCN in Cora and Pudmed, and much worse than HC-GNN-\textit{Louvain}. 
It implies that hierarchical communities generated from the graph topology genuinely lead to a positive effect on information propagation.
Meanwhile, it is surprisingly found that HC-GNN-Random achieves better performance than GCN on Citeseer. We argue this is because HC-GNN-Random has the ability to spread supervision information in the hierarchy structure, leading to the occasional improvement. 
To answer (B), we utilise \textit{Girvan Newman}~\cite{GN02} to produce the hierarchical structure by following the same way described in Sec.~\ref{subsec:hierarchical_message-passing_graph_neural_networks}, and have a model named HC-GNN-\textit{Girvan Newman}.
The results are shown in Fig.~\ref{fig:influence_comparisons}(b).
Although HC-GNN-\textit{Girvan Newman} is not as effective as HC-GNN-\textit{Louvain}, they still outperform GCN.
Such a result indicates that the approaches to generate the hierarchical structure will influence the capability of HC-GNN.
While HC-GNN-\textit{Louvain} leads to promising performance, one can search for a proper hierarchical community detection method to perform better on different tasks.

\begin{figure*}[!ht]
\centering
\includegraphics[width=1.\linewidth]{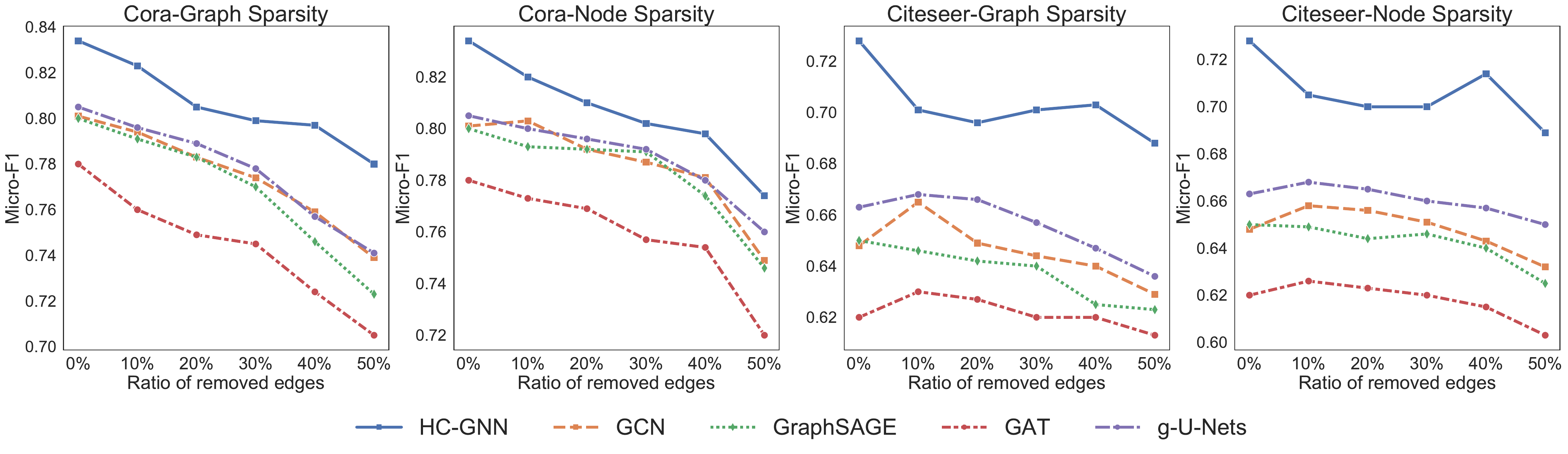}
\caption{ 
Results on semi-supervised node classification in graphs by varying the percentage of removed edges.
}
\label{fig:comparison_sparsity}
\vspace{-3mm}
\end{figure*}

\smallskip\noindent
\textbf{Influence of graph sparsity (RQ5).}
Since community detection algorithms are sensitive to the sparsity of the graph~\cite{NN12},
we aim at studying how HC-GNN perform under graphs with low sparsity values in the task of semi-supervised node classification.
We consider two kinds of sparsity: one is graph sparsity by randomly removing a percentage of edges from all edges in the graph, i.e., $10\%-50\%$; the other is node sparsity by randomly drawing a portion of edges incident to every node in the graph.
The random removal of edges can be considered that users hide partial connections due to privacy concerns.
The results for Cora and Citeseer are presented in Fig.~\ref{fig:comparison_sparsity}.
HC-GNN significantly outperforms the competing methods on graph sparsity and node sparsity under different edge-removal percentages. 
Such results prove that even though communities are subject to sparse graphs, but it will not damage HC-GNN's performance making it worse than other competing models.

\begin{table}[!ht]
\caption{
Comparison of HC-GNN with different primary GNN encoders (\textit{within-level propagation}), follow the transductive node classification settings. 
Reported results in Micro-F1. 
}
\label{table:ablation_analysis_primary_encoder}
\centering
\begin{tabular}{l  c  c  c  c}
\hline
Models & \textbf{Cora}  & \textbf{Citeseer}  & \textbf{Pubmed}     \\
\hline
\hline
GCN             & 0.802 & 0.648 & 0.779 \\
HC-GNN w/ GCN   & \textbf{0.834} & \textbf{0.728} & \textbf{0.812} \\
\hline
GAT             & 0.772 & 0.629 & 0.775 \\
HC-GNN w/ GAT   & \textbf{0.801} & \textbf{0.712} & \textbf{0.819} \\
\hline
GCNII 	        & 0.823 & 0.722 & 0.791 \\
HC-GNN w/ GCNII & \textbf{0.841} & \textbf{0.734} & \textbf{0.816} \\	
\hline
\end{tabular}
\vspace{-3mm}
\end{table}

\smallskip\noindent
\textbf{Ablation study of different primary GNN encoders (RQ6)}.
We adopted GCN as the default primary GNN encoder in model presentation (Sec.~\ref{sec:proposed_approach}) and previous experiments. 
Here, we present more experimental results by endowing HC-GNN with advanced GNN encoders in Table~\ref{table:ablation_analysis_primary_encoder}. 
The table demonstrates that advanced GNN encoders can still benefit from the multi-grained semantics of HC-GNN. 
For instance, GCNII can stack lots of layers to capture long-range information; however, it still follows a \textit{flat} message-passing mechanism hence naturally ignoring the multi-grained semantics.
HC-GNN further ameliorates this problem for better performance. 


\section{Conclusion and Future Work}
\label{sec:conclusion_and_future_work}
This paper has presented a novel \textit{Hierarchical Message-passing Graph Neural Networks} (HMGNNs) framework, which deals with two critical deficiencies of the \textit{flat} message passing mechanism in existing GNN models, i.e., the limited ability for information aggregation over long-range and infeasible in encoding meso- and macro-level graph semantics.
Following this innovative idea, we further presented the first implementation, \textit{Hierarchical Community-aware Graph Neural Network} (HC-GNN), with the assistance of a hierarchical communities detection algorithm.
The theoretical analysis confirms HC-GNN's significant ability in capturing long-range interactions without introducing heavy computation complexity.
Extensive experiments conducted on $9$ datasets show that HC-GNN can consistently outperform state-of-the-art GNN models in $3$ tasks, including node classification, link prediction, and community detection, under settings of transductive, inductive, and few-shot learning. 
Furthermore, the proposed hierarchical message-passing GNN provides model flexibility. 
For instance, it friendly allows different choices and customised designs of the hierarchical structure, and it incorporates well with advanced flat GNN encoders to obtain more impressive results. 
That said, the HMGNNs could be easily applied to work as a general practical framework to boost downstream tasks with arbitrary hierarchical structure and encoder. 

The proposed hierarchical message-passing GNNs provide a good starting point for exploiting graph hierarchy with GNN models. 
In the future, we aim to incorporate the learning of the hierarchical structure into the model optimisation of GNNs such that a better hierarchy can be searched on the fly. 
Moreover, it is also interesting to extend our framework for heterogeneous networks.


\bmhead{Acknowledgments}
This work is supported by the Luxembourg National Research Fund through grant PRIDE15/10621687/SPsquared, and supported by Ministry of Science and Technology (MOST) of Taiwan under grants 110-2221-E-006-136-MY3, 110-2221-E-006-001, and 110-2634-F-002-051.

\bibliography{full_format_references} 

\appendix
\clearpage
\section{Introduction of Community Detection Algorithms}
\label{sec:appendix_introduction_of_community_detection_algorithms}
\subsection{\textit{Louvain} Community Detection Algorithm}
\label{subsec:appendix_louvain_community_detection_algorithm}

This section gives necessary background knowledge about the \textit{Louvain}~\cite{BGLL08} community detection algorithm we used in this paper. 
Generally, this is a method to extract communities from large scale graphs by optimising modularity. 

\noindent
\textbf{Modularity.}
The problem of community detection requires the partition of a network into communities of densely connected nodes, with the nodes belonging to different communities being only sparsely connected.
The so-called modularity of the partition often measures the modularity of the partitions resulting from these methods. 
The modularity of a partition is a scalar value between $-1$ and $1$ that measures the density of links inside communities as compared to links between communities and can be defined as~\cite{BGLL08}:
\begin{equation}
\label{eq:def:modulity}
Q = \frac{1}{2m} \sum_{i,j} \biggl[ A_{ij} - \frac{k_i k_j}{2m} \biggr] \delta(c_i,c_j),
\end{equation}
where $c_i$ is the community to which node~$v_i$ is assigned, $k_i$ and $k_j$ are the sum of weights of the edges attached to nodes $v_i$ and $v_j$, respectively. 
The $\delta$-function $\delta(u,v)$ is $1$ if $u=v$ and $0$ otherwise and $m=\frac{1}{2}\sum_{ij} A_{ij}$.

In order to maximise this value ($Q$) efficiently, the \textit{Louvain} community detection algorithm has two main phases that are repeated iteratively:
\textit{(i)} each node in the graph is assigned to its own community; 
\textit{(ii)} for each node, the change in modularity is calculated by removing $v$ from its own community and moving it into the community of each neighbour $u$ of $v$.
This value is easily calculated by two steps: (1) removing $v$ from its original community and (2) inserting $v$ into the community of $u$. 
This is a typical greedy optimisation, some following work proposed solutions to optimise its efficiency significantly~\cite{BGLL08}.

\subsection{\textit{Girvan Newman} Community Detection Algorithm}
\label{subsec:appendix_girvan_newman_community_detection_algorithm}

The \textit{Girvan-Newman} algorithm \cite{GN02} for the detection and analysis of community structure relies on the iterative elimination of edges that have the highest number of shortest paths between nodes passing through them. 
By removing edges from the graph one by one, the network breaks down into smaller pieces, so-called communities. 

The \textbf{betweenness centrality} \cite{F77} of a node $v$ is defined as the number of shortest paths between pairs of other nodes that run through $v$. 
It is a measure of the influence of a node over the flow of information between other nodes, especially in cases where information flow over a network primarily follows the shortest available path.
Based on the definition of betweenness centrality, the \textit{Girvan-Newman} algorithm can be generally divided into four main steps:
\begin{enumerate}
    \item For every edge in a graph, calculate the edge betweenness centrality.
    \item Remove the edge with the highest betweenness centrality.
    \item Calculate the betweenness centrality for every remaining edge.
    \item Repeat steps $2-3$ until there are no more edges left.
\end{enumerate}


\section{Competing methods}
\label{sec:appendix_competing_methods}
\textbf{Competing methods.}
To validate the effectiveness of HC-GNN, we compare it with $9$ competing methods which include $6$ flat message-passing GNN models,
$2$ hierarchical GNN models 
and another state-of-the-art model.
\begin{itemize}
\item GCN\footnote{\url{https://github.com/tkipf/pygcn}}~\cite{KW17} is the first deep learning model which generalises the convolutional operation on graph data and introduces the semi-supervised train paradigm.
\item GraphSAGE\footnote{\url{https://github.com/williamleif/GraphSAGE}}~\cite{HYL17} extends the convolutional operation of GCN to mean/ max/ LSTM convolutions
and introduces a sampling strategy before employing convolutional operations on neighbour nodes.
\item GAT\footnote{\url{https://github.com/PetarV-/GAT}}~\cite{VCCRLB18} employs trainable attention weight during message aggregation from neighbours,
which makes the information received by each node different and provides interpretable results.
\item GIN\footnote{\url{https://github.com/weihua916/powerful-gnns}}~\cite{XHLJ19} summarises previous existing GNN layers as two components, \textsc{Aggregate} and \textsc{Combine},
and models injective multiset functions for the neighbour aggregation.
\item HARP\footnote{\url{https://github.com/GTmac/HARP}}~\cite{CPHS18} is a hierarchical structure by various collapsing methods for unsupervised node representation learning.
\item P-GNNs\footnote{\url{https://github.com/JiaxuanYou/P-GNN}}~\cite{YYL19} introduces anchor-set sampling to generate node representation with global position-aware.
\item \textsc{g-U-Net}\footnote{\url{https://github.com/HongyangGao/Graph-U-Nets}}~\cite{GJ19} generalises the U-nets architecture of convolutional neural networks for graph data to get better node representation.
It constructs a hierarchical structure with the help of pooling and unpooling operators.
\item GraphRNA\footnote{\url{https://github.com/xhuang31/GraphRNA_KDD19}}~\cite{HSLH19} proposes using recurrent neural networks to capture the long-range node dependencies to assist GNN to obtain better node representation.
\item GXN~\cite{LCZT20}\footnote{\url{https://github.com/limaosen0/GXN}} proposes an infomax pooling operator for graph data to get the hierarchy structure. 
\item GCNII\footnote{\url{https://github.com/chennnM/GCNII}}~\cite{CWHDL20} simplifies the aggregation design of flat GNNs, joined with well-designed normalisation units to get much deeper GNN models. 
\end{itemize}

\section{Model Comparison}
\label{sec:appendix_model_comparison}
\begin{table}[!ht]
\caption{Model comparison in aspects of Node-wise Task (NT), SUPervised training paradigm (SUP), Transductive Inference (TI), Inductive Inference (II), Long-range Information (LI), and Hierarchical Semantics for Node Representations (HSNR). 
}
\label{table:comparison_diff_GNNs}
\centering
\begin{tabular}{l | c | c | c | c | c | c }
\hline
                    & NT        & SUP               & TI             & II         & LI            & HSNR                \\
\hline\hline
GCN~\cite{KW17}        & $\surd$           & $\surd$           & $\surd$       & $\surd$   &               &                   \\
\hline
GraphSAGE~\cite{HYL17} & $\surd$           & $\surd$           & $\surd$       & $\surd$   &               &                   \\
\hline
GAT~\cite{VCCRLB18}    & $\surd$           & $\surd$           & $\surd$       & $\surd$   &               &                   \\
\hline
GIN~\cite{XHLJ19}      & $\surd$           & $\surd$           & $\surd$       & $\surd$   &               &                   \\
\hline
P-GNNs~\cite{YYL19}    & $\surd$           & $\surd$           & $\surd$       &           & $\surd$       &                   \\
\hline
GCNII~\cite{CWHDL20}    & $\surd$           & $\surd$           & $\surd$       & $\surd$          & $\surd$       &                   \\
\hline
\textsc{DiffPool}~\cite{YYMRHL18} &           & $\surd$          & $\surd$       & $\surd$   &               &           \\
\hline
\textsc{g-U-Net}~\cite{GJ19}   & $\surd$           & $\surd$           & $\surd$       & $\surd$   &               & $\surd$            \\
\hline 
\textsc{AttPool}~\cite{HLLLL19}   &           & $\surd$           & $\surd$       & $\surd$   &               &           \\
\hline
ASAP~\cite{RST20}   &           & $\surd$           & $\surd$       & $\surd$   &               &           \\
\hline
GXN~\cite{LCZT20}   & $\surd$   & $\surd$           & $\surd$       & $\surd$   &               &  $\surd$         \\
\hline
GraphRNA~\cite{HSLH19} & $\surd$           & $\surd$           & $\surd$       &           &               &                   \\
\hline
HARP~\cite{CPHS18}     & $\surd$        &         & $\surd$       &           &               &           \\
\hline
LouvainNE~\cite{BMDGM20}& $\surd$       &       & $\surd$       &           &               &           \\
\hline
HC-GNN                 & $\surd$           & $\surd$           & $\surd$       & $\surd$   & $\surd$       & $\surd$           \\
\hline
\end{tabular}
\end{table}

In Section~\ref{sec:related_work}, we have systematically discussed related work and highlighted the differences between HC-GNN and them. 
Here, we further present Table~\ref{table:comparison_diff_GNNs} to summarise the critical advantages of the proposed HC-GNN and compare it with a number of state-of-the-art methods published recently.
We are the first to present the hierarchical message passing to efficiently model long-range informative interaction and multi-grained semantics. 
In addition, our HC-GNN can utilise the community structures and be applied for transductive, inductive and few-shot inferences.




\end{document}